\pdfoutput=1
\documentclass[sigconf]{acmart}
\graphicspath{ {../figs/} }

\AtBeginDocument{%
  \providecommand\BibTeX{{%
    \normalfont B\kern-0.5em{\scshape i\kern-0.25em b}\kern-0.8em\TeX}}}

\copyrightyear{2022}
\acmYear{2022}
\setcopyright{acmcopyright}\acmConference[KDD '22]{Proceedings of the 28th ACM SIGKDD Conference on Knowledge Discovery and Data Mining}{August 14--18, 2022}{Washington, DC, USA}
\acmBooktitle{Proceedings of the 28th ACM SIGKDD Conference on Knowledge Discovery and Data Mining (KDD '22), August 14--18, 2022, Washington, DC, USA}
\acmPrice{15.00}
\acmDOI{10.1145/3534678.3539376}
\acmISBN{978-1-4503-9385-0/22/08}

\usepackage{subfloat}
\usepackage{amsfonts}
\usepackage{amsmath}
\usepackage{booktabs}
\usepackage{multirow}
\usepackage{multicol}
\usepackage{comment}
\usepackage{algpseudocode}
\usepackage{algorithm,algorithmicx}
\usepackage{amsthm}
\usepackage{color}
\usepackage[shortlabels]{enumitem}
\usepackage{subfigure}
\usepackage{graphicx}
\usepackage{threeparttable}

\newcommand{\trans}{^\top}

\newcommand{\tr}{\mathrm{Tr}}

\newcommand{\hsic}{\mathrm{HSIC}}
\newcommand{\coco}{\mathrm{COCO}}

\newcommand{\RR}{\mathbb{R}}
\newcommand{\bb}[1]{\mathbf{#1}}

\newcommand{\dx}{d_X}
\newcommand{\dy}{k}

\newcommand{\nol}{M}
\newcommand{\loss}{\mathcal{L}}
\newcommand{\rloss}{\tilde{\loss}}

\DeclareMathOperator{\E}{\mathbb{E}}

\begin{document}

\title{Bilateral Dependency Optimization: Defending Against Model-inversion Attacks}

\author{Xiong Peng}
\authornote{Equal contribution.}

\affiliation{
  \institution{Hong Kong Baptist University}
  \city{Hong Kong SAR}
  \country{China}
}
\email{xpeng1997@gmail.com}

\author{Feng Liu}
\authornotemark[1]
\affiliation{
  \institution{The University of Melbourne}
  \city{Melbourne}
  \country{Australia}
}
\email{fengliu.ml@gmail.com}

\author{Jingfeng Zhang}
\affiliation{
  \institution{RIKEN AIP}
  \city{Tokyo}
  \country{Japan}
}
\email{jingfeng.zhang@riken.jp}

\author{Long Lan}
\authornote{Corresponding author.}

\affiliation{
    \institution{Hong Kong Baptist University}
    \city{Hong Kong SAR}
    \country{China}
}
\email{longoahead@gmail.com}

\author{Junjie Ye}
\affiliation{
    \institution{The Hong Kong Polytechnic University}
    \city{Hong Kong SAR}
    \country{China}
}
\email{kourenmu@gmail.com}

\author{Tongliang Liu}
\affiliation{
  \institution{The University of Sydney}
  \city{Sydney}
  \country{Australia}
}
\email{tongliang.liu@sydney.edu.au}

\author{Bo Han}
\authornotemark[2]

\affiliation{
  \institution{Hong Kong Baptist University}
  \city{Hong Kong SAR}
  \country{China}
}
\email{bhanml@comp.hkbu.edu.hk}

\renewcommand{\shortauthors}{Xiong Peng et al.}

\begin{abstract}

Through using only a well-trained classifier, \emph{model-inversion} (MI) attacks can recover the data used for training the classifier, leading to the privacy leakage of the training data. 
To defend against MI attacks, previous work utilizes a \emph{unilateral} dependency optimization strategy, i.e., minimizing the dependency between inputs (i.e., features) and outputs (i.e., labels) during training the classifier. 
However, such a minimization process conflicts with minimizing the supervised loss that aims to maximize the dependency between inputs and outputs, causing an \emph{explicit} trade-off between model robustness against MI attacks and model utility on classification tasks. 
In this paper, we aim to minimize the dependency between the latent representations and the inputs while maximizing the dependency between latent representations and the outputs, named a \emph{bilateral dependency optimization} (BiDO) strategy.
In particular, we use the dependency constraints as a universally applicable regularizer in addition to commonly used losses for deep neural networks (e.g., cross-entropy), which can be instantiated with appropriate dependency criteria according to different tasks. 
To verify the efficacy of our strategy, we propose two implementations of BiDO, by using two different dependency measures: \emph{BiDO with constrained covariance} (BiDO-COCO) and \emph{BiDO with Hilbert-Schmidt Independence Criterion} (BiDO-HSIC). 
Experiments show that BiDO achieves the state-of-the-art defense performance for a variety of datasets, classifiers, and MI attacks while suffering a minor classification-accuracy drop compared to the well-trained classifier with no defense, which lights up a novel road to defend against MI attacks.

\end{abstract}



\maketitle

\section{Introduction}
\label{sec:introduction}

\begin{figure}[!t]
\centering
    \vspace{10pt}
    \subfigure[$\lambda=1; \lambda=5$]{
    \includegraphics[width=0.22\textwidth]{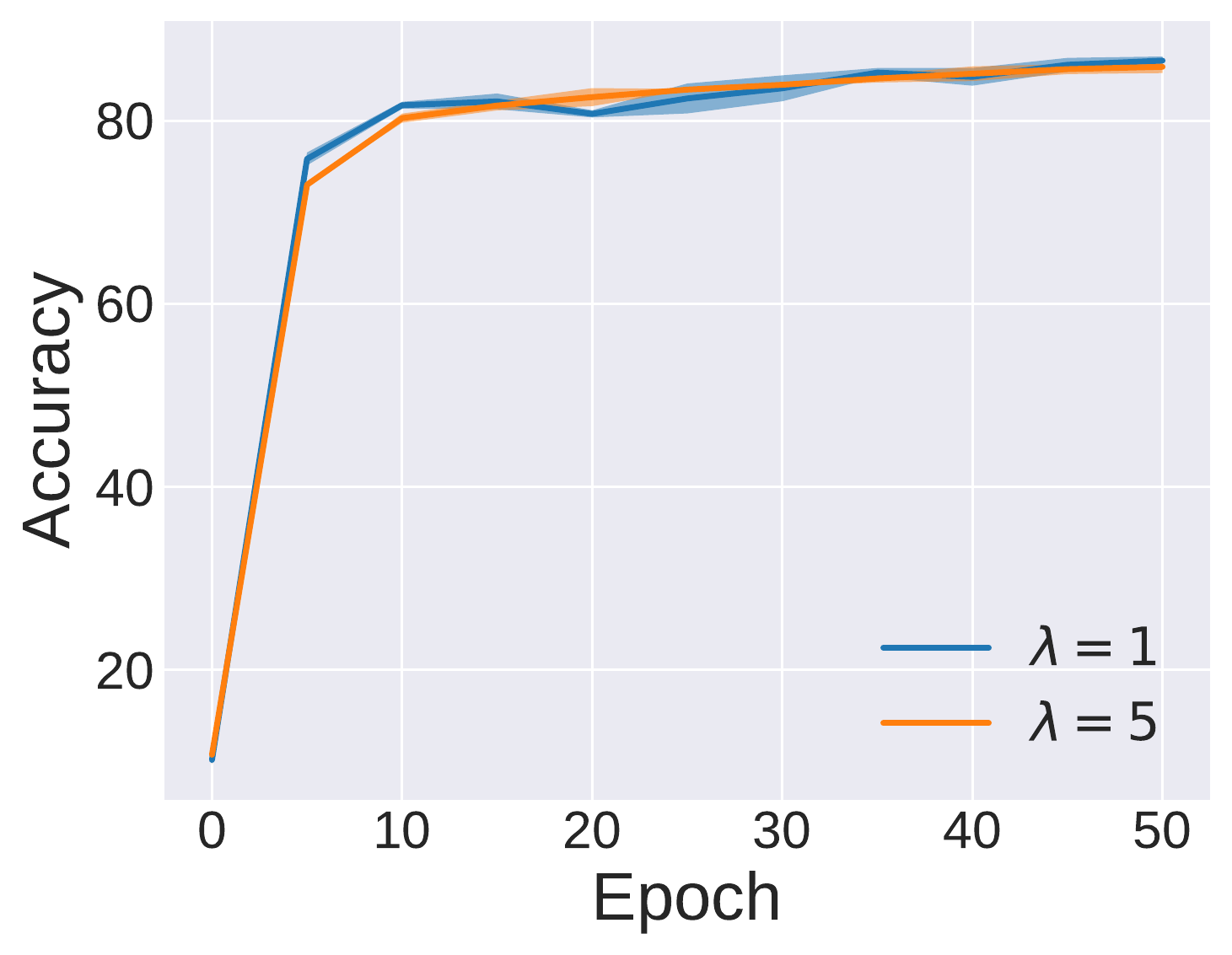}
    }
    \subfigure[$\lambda=20; \lambda=100$]{
    \includegraphics[width=0.228\textwidth]{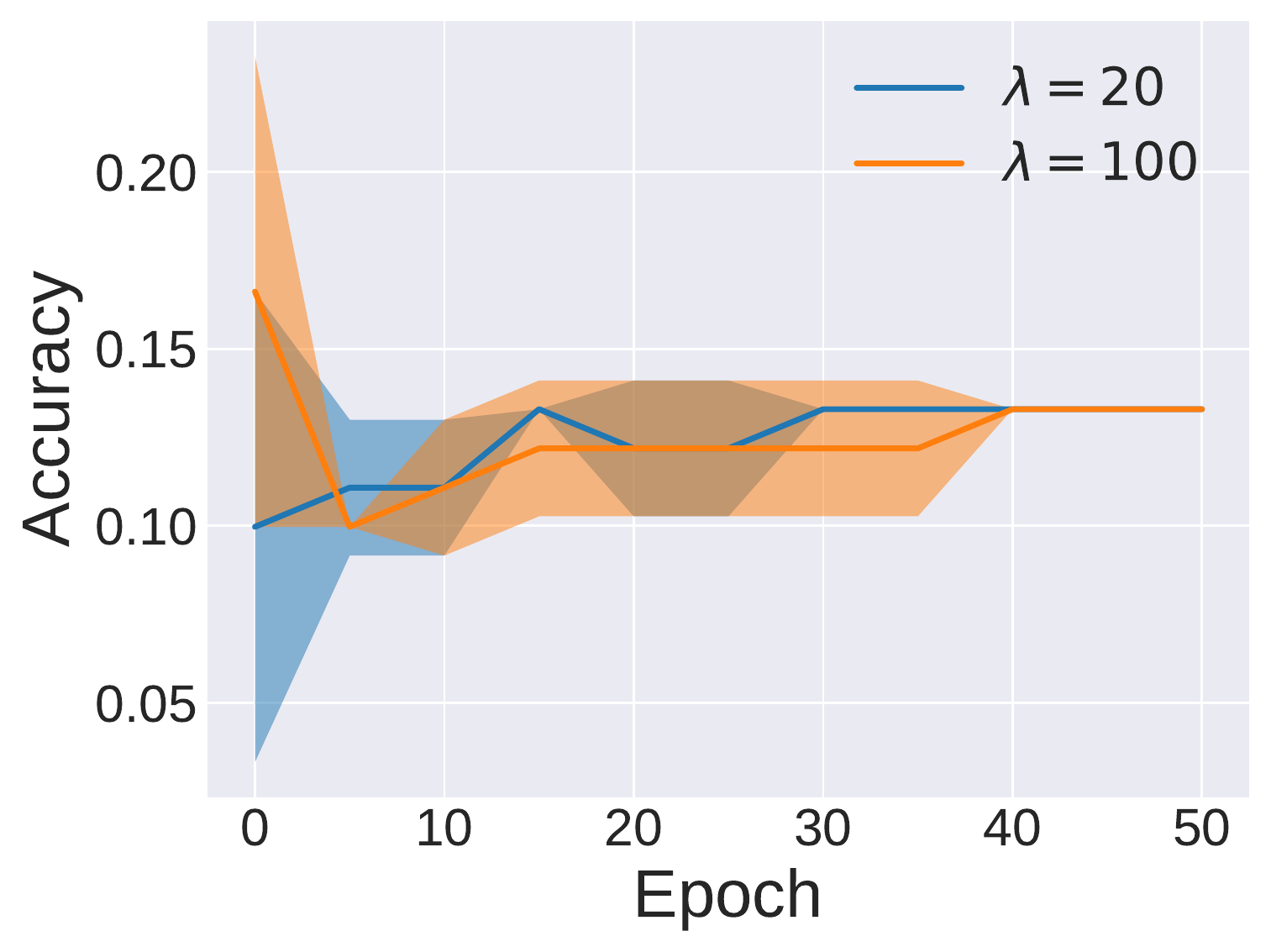}
    }

    \vspace{-10pt}
	\caption{Minimizing the supervised loss conflicts with minimizing the dependency between inputs and outputs. These figures illustrate the test accuracy of a face recognition model during training, which incorporates $d(X, \hat{Y})$, the dependency between inputs and outputs, into the training objective as a regularizer (the previous method to defend against model-inversion attacks \cite{MID}). We use \emph{constrained covariance} as the dependency measure $d(X, \hat{Y})$, $\lambda$ denotes the balancing hyper-parameter of the regularization term. 
	It is clear that the previous method causes an explicit accuracy drop (from $\sim 85\%$ (a) to $\sim 0.15\%$ (b)) when we increase $\lambda$ beyond some threshold (from $\lambda=5$ (a) to $\lambda=20$ (b)).}
	\label{fig:motivation}
	\vspace{-1.5em}
\end{figure}

With the development of \emph{Machine learning} (ML) algorithms, \emph{deep neural networks} (DNNs)~\cite{vgg16,resnet34,transformer}
are increasingly adopted in various privacy-sensitive applications, such as facial recognition~\cite{facenet,taigman2014deepface}, medical diagnoses~\cite{ma2017dipole_medical}, and intelligent virtual assistants~\cite{VA}. Since training DNNs could involve processing sensitive and proprietary datasets in privacy-related applications, there are great concerns about privacy leakage. To protect the privacy of individuals whose personal information is used during the training, enterprises typically release only well-trained DNNs through ML-as-a-services platforms, wherein users can download pre-trained models (e.g., Pytorch Hub) or query the model via some sort of programming or user interfaces (e.g., Amazon Recognition), which are referred to as white-box access and black-box access, respectively. 

Unfortunately, the white-box access or black-box access can still cause privacy leakage. Recent studies have revealed that DNNs may expose sensitive information in the training data under various privacy attacks~\cite{privacy_attacks}. For example, in the growing paradigm of federated learning~\cite{li2021fedrs,fair_FL}, wherein a malicious collaborator could infer private information of other data centers through their uploaded model updates (i.e., gradients)~\cite{huang2021evaluating_g}. More importantly, by using only a well-trained classifier (i.e., a target model), \emph{model-inversion} (MI) attacks can recover the data used for training the classifier, leading to the privacy leakage of the training data. 

The first MI attack~\cite{First_MI} has been proposed to predict genetic markers which are used as part of the input features, where an adversary was given the model and other possible auxiliary information. In recent works~\cite{GMI,DMI,VMI}, MI attacks have been widely adopted to recover facial images of any person from a well-trained face recognition model, once the adversary has successfully reconstructed the facial images of individuals in the private training set, they could abuse them to break into otherwise secure systems.

Since the MI adversary exploits the correlation between model inputs and outputs for successful attacks, a recent work~\cite{MID} named \emph{mutual information regularization based defense} (MID) (see Figure~\ref{fig:framework}(a)) utilizes a \emph{unilateral} dependency optimization strategy, i.e., minimizing the dependency between inputs and outputs during training the model. Specifically, they introduce a mutual information regularizer into the training objective, which penalizes the mutual information between model inputs and outputs. 

However, such a minimization process conflicts with minimizing the supervised loss that aims to maximize the dependency between inputs and outputs, causing an \emph{explicit} trade-off between model robustness against MI attacks and model utility on classification tasks. We further conducted experiments to verify this claim, as illustrated in Figure~\ref{fig:motivation}, once the balancing hyper-parameter $\lambda$ of the dependency term crosses some threshold, the model cannot learn anything from the training set. Besides, since mutual information is notoriously expensive to compute, MID resorted to the variational approximations of the mutual information term rather than the actual quantity. 

In this paper, we propose to minimize the dependency between the latent representations and the inputs while maximizing the dependency between latent representations and the outputs, named a \emph{bilateral dependency optimization} (BiDO) strategy (see Figure~\ref{fig:framework}(b)). The former constraint limits redundant information propagated from the inputs to the latent representations which can be exploited by the adversary, thus improving the model's ability to prevent privacy leakage; while the latter drives the latent layers to learn discriminative representations which ensure model utility on classification tasks. 

Specifically, we use the dependency constraints as a universally applicable regularizer in addition to commonly used losses for DNNs (e.g., cross-entropy), which can be realized with appropriate dependence criteria according to different tasks. To verify the efficacy of our strategy, we propose two implementations of BiDO, by using two different dependency measures: \emph{BiDO with constrained covariance}~\cite{COCO} (BiDO-COCO) and \emph{BiDO with Hilbert-Schmidt independence criterion}~\cite{HSIC} (BiDO-HSIC).

Experiments show that BiDO achieves the state-of-the-art defense performance for a variety of datasets (including CelebA~\cite{celeba}, MNIST~\cite{MNIST} and CIFAR-10~\cite{cifar10}), target models (including LeNet~\cite{MNIST}, VGG16~\cite{vgg16}, and  ResNet-34~\cite{resnet34}) and MI attacks (including \emph{generative MI attack} (GMI) \cite{GMI}, \emph{knowledge-enriched distributional MI attack} (KED-MI) \cite{DMI} and \emph{variational MI attack} (VMI) \cite{VMI}), while suffering a minor classification-accuracy drop compared to the well-trained classifier with no defense.  As a highlight, when we evaluate the defense performance on CelebA against KED-MI, BiDO-HSIC reduces attack accuracy for more than $29\%$, top-5 attack accuracy for more than $20\%$ while only suffering $\sim 6\%$ classification accuracy decline (see Table~\ref{tbl:DMI_celeba}). These empirical results show the effectiveness of BiDO and light up a novel road to defend against MI attacks.


\section{Related Works}
\label{sec:background}
In this section, we briefly review MI attacks, the defense strategies against MI attacks, and dependency measures that are used to see how two random variables are dependent.

\subsection{Model-inversion Attacks}
The general goal of privacy attacks against the ML model is to extract information about the training data or to extract the model itself (model extraction attacks~\cite{model_extraction}). The attacks related to exploiting training data information can be further categorized into membership inference attacks~\cite{label-only}, model inversion attacks~\cite{GMI,DMI,VMI,First_MI}, and property inference attacks~\cite{property_inference}, each with a different specific goal. In this paper, we focus on the defense mechanism against MI attacks, which aims at inferring sensitive information in the training set given only access to a well-trained target model.

The first MI attack was proposed in the context of genomic privacy~\cite{First_MI}, where the authors try to infer private genomic attributes about individuals given the linear regression model that uses them as input features, the response of the model, as well as other non-sensitive features of the input. Algorithmically, they formulated the MI attack as an optimization problem in the input space, seeking for sensitive features that achieve the maximum likelihood under the target model. Fredrikson et al.~\cite{Second_MI} extended the attack algorithm to more challenging tasks, e.g., recovering facial images of a person from a face recognition model, and more complex target models, e.g., decision trees and shallow neural networks. 

However, when the attack scenarios involve recovering private data which lies in high-dimensional and continuous data spaces (e.g., image-based spaces), their attack algorithm failed, since directly optimizing over the input space could result in imperceptible and unnatural features. To handle this problem, MI attacks based on deep generators~\cite{DCGAN,styleGAN} were proposed~\cite{GMI,DMI,VMI,Knowledge_Align}. GMI~\cite{GMI} proposed a two-stage attack framework, which first pre-trains \emph{generative adversarial networks} (GANs)~\cite{DCGAN} on a public auxiliary dataset to distill generic prior, and then uses it to guide the gradient-based attack in the latent space of the generator. 

KED-MI~\cite{DMI} addressed two limitations in GMI. First, they leverage the target model to label the public dataset and train the discriminator to differentiate not only the real and fake samples but also the labels, which enables the generator to distill the private knowledge customized for the specific classes of interest in the target model; second, instead of generating a representative image for a given label, they propose to explicitly parameterize the private data distribution and solve the attack optimization over the distributional parameters. VMI~\cite{VMI} views the MI attack problem as a variational inference problem and provides a unified framework for analyzing existing MI attacks from a Bayesian probabilistic perspective. 

Moreover, Yang et al.~\cite{Knowledge_Align} studied the black-box MI attack and trained a separate generator that swaps inputs and outputs of the target model, using an architecture of autoencoder. Salem et al.~\cite{updates_leak} studied the black-box MI attack for online learning, where the adversary has access to the versions of the target model before and after an online update and the goal is to recover the training data used to perform the update.

\subsection{Defending against MI Attacks}
Improving model robustness against MI attacks is critical to the privacy protection of training data, however, the research on such defense mechanisms is still limited. \emph{Differential privacy} (DP) is a widely adopted privacy-protection technique whose effectiveness is theoretically guaranteed. In~\cite{First_MI}, DP was used to add noise to the objective function of the optimization problem, while in~\cite{GMI} DP was used to add noise to the gradient during the optimization of the model. Despite the rigorous theoretical privacy guarantee of DP, the aforementioned works~\cite{GMI,First_MI} have experimentally shown that \emph{current} DP mechanisms do not mitigate the MI attacks while retaining desirable model utility at the same time.

In a recent work, Wang et al.~\cite{MID} introduced \emph{mutual information regularization based defense} (MID) against MI attacks, which also provides insights into the empirical inefficacy of differential privacy techniques from a theoretical viewpoint. They propose to limit the dependency between the model inputs and outputs directly by introducing a mutual information regularizer into the training objective, which penalizes the mutual information between the inputs and outputs during the training process. However, since mutual information is unfortunately intractable in continuous and high-dimensional settings, MID resorted to mutual information approximations rather than the actual quantity. Moreover, though limiting correlation between the model inputs and outputs seems intuitive to defend against MI attacks, such a minimization process conflicts with minimizing the supervised loss that aims to maximize the dependency between inputs and outputs, causing an \emph{explicit} trade-off between model robustness against MI attacks and model utility on classification tasks. 

There are also defense mechanisms for black-box attacks that are relatively easy to be defended. In this setting, the defender can only modify output confidence scores to weaken the correlation between the inputs and outputs. Such a modification could be injecting noise into the confidence scores~\cite{updates_leak}, reducing their precision~\cite{First_MI}, or purifying the distinguishable pattern among them~\cite{purifier}.

\subsection{Statistical Dependency Measures}
Measuring dependency among random variables is well established in statistics. Well-known measures include constrained covariance \cite{COCO}, kernel canonical correlation \cite{KCC}, and Hilbert-Schmidt independency criterion \cite{HSIC}. Recently, researchers have found various applications of these measures, where they are utilized to force DNNs to learn discriminative features. Examples include self-supervised learning of image representations~\cite{SSL_HSIC}, causal representation learning~\cite{causal_repl}, clustering algorithms~\cite{clustering} and feature selection~\cite{feature_selection}.

\section{preliminaries}
\label{sec:preliminary}

In this section, we present the definitions of two statistical dependency measures that we used to realize the BiDO defense mechanism, namely the \emph{constrained covariance} (COCO)~\cite{COCO} and the  \emph{Hilbert-Schmidt independency criterion} (HSIC)~\cite{HSIC}. 

\subsection{Constrained Covariance (COCO)}
COCO is a kernel-based dependency measure between random variables. Given function classes $\mathcal F, \mathcal G$ containing subspaces $F \in \mathcal{F}$ and $G \in \mathcal{G}$, and a probability measure $\mathbb{P}_{XY}$, COCO is defined as:
\begin{equation}
    \coco(X, Y) = \sup_{f \in  F, g \in G} \left[\E(f(X)g(Y)) - \E(f(X))\E(g(Y)) \right].
\end{equation}
For the sets of continuous functions $\mathcal{F}$ and $\mathcal{G}$ that are bounded by $1$, $\coco(X, Y)=0$ if and only if $X$ and $Y$ are independent~\cite{COCO}, and larger values of the measure correspond to stronger dependency.

Given samples
$\{(x_i, y_i)\}_{i= 1}^{N}$
drawn i.i.d.\ from $\mathbb{P}_{XY}$, the empirical estimate of COCO, $\widehat\coco(X, Y)$, is defined as:
\begin{equation*}
\small
    \sup_{f \in  F, g \in G} \left[\frac{1}{N}\sum_{i=1}^N{f(x_i)g(y_i)} - \frac{1}{N^2}\sum_{i=1}^N{f(x_i)\sum_{j=1}^N{g(y_j)}} \right].
\end{equation*}

When $\mathcal{F}$ and $\mathcal{G}$ are \emph{reproducing kernel Hilbert spaces} (RKHS \cite{gretton2012kernel,liu2020learning,liu2021meta}), inner products in an RKHS are by definition \emph{kernel functions}: 
$k(x_i,x_j) = \left\langle \phi(x_i), \phi(x_j) \right\rangle_{\mathcal{F}}$ 
and 
$l(y_i,y_j) = \left\langle \psi(y_i), \psi(y_j) \right\rangle_{\mathcal{G}}$. 
Given $F$ and $G$ their respective unit balls, then
\begin{equation}\label{eq:empirical_coco}
    \widehat\coco(X, Y) = \frac{1}{N} \sqrt{\left\lVert \widetilde{K}\widetilde{L} \right\rVert_2},
\end{equation}
where $\widetilde{K}=HKH$ and $\widetilde{L}=HLH$, $K_{ij}=k(x_i, x_j)$ and $L_{ij}=l(y_i, y_j)$ are the kernel matrices,
and $H = I - \frac{1}{N} \bb 1 \bb 1\trans$ is the centering matrix.

\subsection{Hilbert-Schmidt Independency Criterion (HSIC)}
HSIC is another kernel-based measure which improved from COCO. For a universal kernel, $\hsic(X, Y) = 0$ if and only if $X$ and $Y$ are independent. The larger value of HSIC represents the stronger dependency.
HSIC measures the dependency between $X$ and $Y$ by first taking a nonlinear feature transformation of each,
say $\phi : \mathcal X \to \mathcal F$
and $\psi : \mathcal Y \to \mathcal G$
(with $\mathcal F$ and $\mathcal G$ RKHS),
and then evaluating the norm of the cross-covariance between those features:
\begin{gather}
    \hsic(X, Y) = \left\lVert
    \E[\phi(X) \, \psi(Y)\trans]
    - \E[\phi(X)] \, \E[\psi(Y)]\trans
    \right\rVert^2_\mathit{HS}
.\end{gather}
Here $\lVert \cdot \rVert_\mathit{HS}$ is the Hilbert-Schmidt norm, which in finite dimensions is the usual Frobenius norm. HSIC measures the scale of the correlation in these nonlinear features, which allows it to identify nonlinear dependencies between $X$ and $Y$ with appropriate features $\phi$ and $\psi$.
Let $(X',Y')$, $(X'', Y'')$ be independent copies of $(X, Y)$, we can compute the HSIC value using the following equation.
\begin{equation} \label{eq:hsic-pop}
        \begin{split}
            \hsic(X, Y) = 
            & \E\left[k(X,X') l(Y,Y')\right] - 2\E\left[ k(X,X') l(Y, Y'') \right] \\
                          & + \E\left[ k(X, X') \right] \E\left[l(Y, Y')\right].
        \end{split}
\end{equation}
HSIC is also straightforward to estimate: given samples
$\{(x_i, y_i)\}_{i= 1}^{N}$
drawn i.i.d.\ from the joint distribution of $(X, Y)$, Gretton et al.~\cite{HSIC} propose an estimator
\begin{align}
    \widehat\hsic(X, Y) &= \frac{1}{(N-1)^2} \tr(KHLH)\,,
    \label{eq:empirical_hsic}
\end{align}
where $K_{ij}=k(x_i, x_j)$ and $L_{ij}=l(y_i, y_j)$ are the kernel matrices,
and $H$ is the centering matrix.

\section{Defense via Bilateral Dependency Optimization}

This section shows our defense strategy against MI attacks. First, we will introduce some key concepts in defending against MI attacks.

\begin{figure*}[t]
  \includegraphics[width=1.0\textwidth]{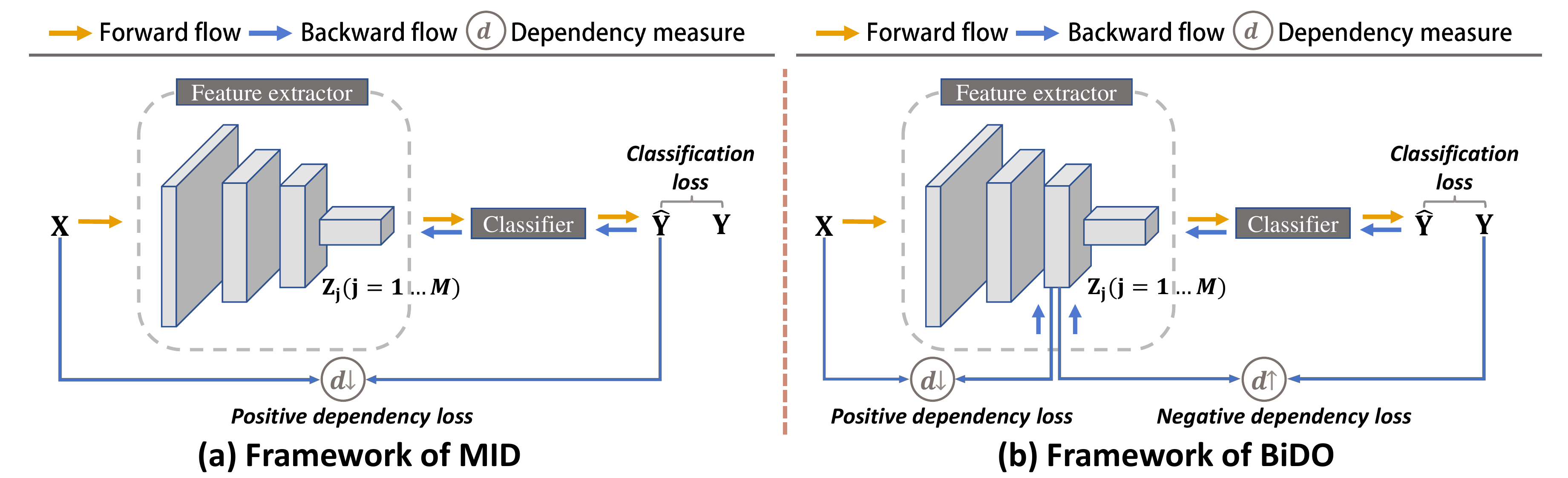}
  \vspace{-20pt}
  \caption{Overview of MID framework vs. \emph{bilateral dependency optimization} (BiDO) framework. BiDO forces DNNs to learn robust latent representations by minimizing $d(X, Z_j)$ to limit redundant information propagated from the inputs to the latent representations while maximizing $d(Z_j, Y)$ to keep the latent representations informative enough of the label.}
  \label{fig:framework}
\end{figure*}

\subsection{Key Concepts}
\paragraph{\bf Model Owner \& Target Classifier}
Given a training dataset $\mathcal{D}_{\rm tr} = \{(x_i, y_i)\}_{i= 1}^{N}$, where $x_i \in \RR^{\dx}, y_i \in \{0,1\}^{\dy}$ are i.i.d.~samples drawn from joint distribution $P_{XY}$. The model owner trains a $\dy$-way classifier $f_\theta:\RR^{\dx} \to \RR^{\dy}$ parameterized by weights $\theta \in \RR^{d_{\theta}}$. 

The model owner has a validation dataset $\mathcal{D}_{\rm val}$ to test the classifier performance, which is drawn from the same underlying private data distribution as $\mathcal{D}_{\rm tr}$. The goal of $f_\theta$ is to make accurate inference on the unseeing dataset $\mathcal{D}_{\rm te}$. The classifier is trained to predict $Y$ from $X$ by optimizing:
\begin{align}
\loss(\theta)= \mathbb{E}_{XY}[\ell(f_{\theta}(X), Y)] \approx \frac{1}{N} \sum_{i=1}^N \ell(f_{\theta}(x_i), y_i)
, \label{eq:loss}
\end{align}
where $\ell:\RR^k \times \RR^k \to \RR$ is a loss function, e.g., cross-entropy. The training process usually terminates when the classification accuracy achieves the best on $\mathcal{D}_{\rm val}$. The model owner releases the well-trained classifier $f_{\theta}$ for downloading, or as a black-box interface. As for the latter setting, the users can only query $f_{\theta}$ with their local data $x \in \mathcal{D}_{\rm te}$ and receive a prediction score vector $\hat{y}$ from $f_{\theta}$, which is probability distribution of the classifier's confidence over all possible classes, and the $i$-th element ${f_{\theta}}^i(x)$ is the probability of the data $x$ belonging to class $i$.

\paragraph{\bf Adversary}
Given access to the "target classifier" $f_{\theta}$, the adversary aims at inferring private information about the training data. In early works~\cite{First_MI,Second_MI,GMI}, the adversary tried to recover the exact instances in the training set. For example, for a class of label $i$, the following optimization problem is solved to synthesize the corresponding input: $\max_{x}{\log f_{\theta}}^i(x)$. However, in real scenarios, a given label should naturally correspond to a distribution of training samples, therefore the most recent works~\cite{DMI,VMI} propose to recover a data distribution w.r.t. label $i$ in the MI attack, which should approximate the class conditional distribution $P(X|Y=i)$.

\paragraph{\bf Defense}
The general goal of defending against MI attacks is to design a strategy to train the target classifier $f_{\theta}$ on $\mathcal{D}_{\rm tr}$ such that the access to the well-trained classifier does not leak much information about the class conditional distribution $P(X|Y=i)$  w.r.t. some specific label $i$.

\subsection{Bilateral Dependency Optimization}

In this section, we will introduce our \emph{bilateral dependency optimization} (BiDO) strategy. The overview of our defense strategy is illustrated in Figure~\ref{fig:framework}.
Given a feedforward DNN $f_{\theta}:\RR^{d_{X}} \to \RR^{k}$ parameterized by $\theta$ with $\nol$ latent layers, and an input $X$, we denoted by $Z_j \in \RR^{d_{Z_j}}$, $j \in \{1, \dots, \nol\}$ the $j$-th latent representation under input $X$. We define the training objective of BiDO with a universally applicable regularizer as follows:
\begin{align}
    \label{eq:obj}
    \begin{split}
    \rloss(\theta) =  \loss(\theta)
        + \lambda_{x} &\sum_{j=1}^{\nol} d(X, Z_j) -\lambda_{y} \sum_{j=1}^{\nol} d(Z_j, Y), 
    \end{split}
\end{align}
where $\loss$ is the standard loss given by Eq.~\eqref{eq:loss}, $d$ is a dependency measure and $\lambda_{x}, \lambda_{y}\in \RR_+$ are balancing hyper-parameters. Together, the second and third terms in Eq.~\eqref{eq:obj} form the BiDO strategy. The former constraint is proposed to limit redundant information propagated from the inputs to the latent representations which could be exploited by the MI adversary for successful attacking, thus improving the model's ability to prevent privacy leakage. Moreover, note that minimizing $d(X, Z_j)$ alone would also lead to the loss of useful information, so it is necessary to keep the $d(Z_j, Y)$ term to make sure $Z_j$ is informative enough of $Y$ as well as to maintain the discriminative nature of the classifier.

\begin{algorithm}[!t]
\caption{Train Robust Model against MI Attacks with BiDO}
\begin{flushleft}
{\bf Input:} input samples $\{(x_i, y_i)\}_{i=1}^{N}$, a classifier $f_{\theta}$ parameterized by $\theta$, dependency measure $d$ (e.g., HSIC), predefined elements needed to compute the dependency (in HSIC, such elements are kernel functions $k_x, k_y, k_z$.), mini-batch size $m$, learning rate $\alpha$, total epoch $T_{max}$.\\
{\bf Output:} a well-trained classifier $f_{\theta}$
\end{flushleft}
\begin{algorithmic}[1]
\State Initialize $t \leftarrow 0$.
\While{$t < T_{max}$} 
    \State {\bfseries Sample} a mini-batch of size $m$ from input samples;
    \State {\bfseries Forward Propagation:} calculate the latent representation $z_j$($j\in{1 \dots M}$), and the output $f_\theta(x)$;
    \State {\bfseries Compute} value of the predefined elements associated with measure $d$; (in HSIC, compute kernel matrices for $X$, $Y$ and $Z_j$ using $k_x, k_y, k_z$ respectively inside a mini-batch.) 
    \State {\bfseries Compute} $\rloss(\theta)$ via Eq.~\eqref{eq:obj};
    \State {\bfseries Backward Propagation:} $\theta \leftarrow \theta - \alpha \nabla \rloss(\theta)$;
    \State $t \leftarrow t+1$;
\EndWhile
\State \Return $f_{\theta}$
\end{algorithmic}
\label{alg:Uinterface}
\end{algorithm}

\subsection{Realizations of BiDO}

The BiDO strategy provides an appealing defense principle. Next, we present two concrete realizations of the universally applicable regularizer, namely BiDO-COCO and BiDO-HSIC.

When we use COCO as the dependency measure, the training objective in Eq.~\eqref{eq:obj} is realized as:
\begin{align}
    \label{eq:coco_obj}
    \begin{split}
    \rloss_{\mathrm{COCO}}(\theta) =  \loss(\theta)
        + \lambda_{x} &\sum_{j=1}^{\nol} \coco(X, Z_j) -\lambda_{y} \sum_{j=1}^{\nol} \coco(Z_j, Y)\,.
    \end{split}
\end{align}
Following the COCO estimator in Eq.~\eqref{eq:empirical_coco}, we have
\begin{subequations}
\begin{align}
    &\coco(X, Z_j) = \frac{1}{N} \sqrt{\left\lVert \widetilde{K}_X\widetilde{K}_{Z_j} \right\rVert_2}\,,
    \hspace{35pt} \\
    &\coco(Z_j, Y) = \frac{1}{N} \sqrt{\left\lVert \widetilde{K}_{Z_j}\widetilde{K}_Y \right\rVert_2}\,.
\end{align}
\end{subequations}

Similarly, when we use HSIC as the dependency measure, the training objective in Eq.~\eqref{eq:obj} is realized as:
\begin{align}
    \label{eq:hsic_obj}
    \begin{split}
    \rloss_{\mathrm{HSIC}}(\theta) =  \loss(\theta)
        + \lambda_{x} &\sum_{j=1}^{\nol} \hsic(X, Z_j) -\lambda_{y} \sum_{j=1}^{\nol} \hsic(Z_j, Y). 
    \end{split}
\end{align}
Following the HSIC estimator in Eq.~\eqref{eq:empirical_hsic}, the HSIC of each term is:
\begin{subequations}
\begin{align}
    &\hsic(X, Z_j) = \frac{1}{(N-1)^2} \tr(K_XHK_{Z_j}H)\,,
    \hspace{35pt} \\
    &\hsic(Z_j, Y) = \frac{1}{(N-1)^2} \tr(K_{Z_j}HK_YH)\,.
\end{align}
\end{subequations}

We present the overall algorithm in Algorithm~\ref{alg:Uinterface}. In the given realizations, we perform SGD over $\rloss_{\mathrm{COCO}}$/$\rloss_{\mathrm{HSIC}}$: both $\loss$ and COCO/HSIC can be evaluated empirically over mini-batches. For the latter, we use the estimator Eq.~\eqref{eq:empirical_coco}/Eq.~\eqref{eq:empirical_hsic}, restricted over the current batch. As we have $m$ samples in a mini-batch, the complexity of calculating the empirical COCO/HSIC is $O(m^3)$/$O(m^2)$ for a single layer. Thus, the overall complexity for Eq.~\eqref{eq:coco_obj}/Eq.~\eqref{eq:hsic_obj} is $O(Mm^3)$/$O(Mm^2)$~\cite{HSIC}. This computation is highly parallelizable, thus, the additional computation time of BiDO-COCO or BiDO-HSIC is small when compared to training a neural network via cross-entropy only. Note that, BiDO-COCO and BiDO-HSIC are only shown as examples of how to implement BiDO. We can also implement BiDO with other dependency measures, such as kernel canonical correlation \cite{KCC}.

\section{Experiments}
In this section, we present the experimental evaluation for verifying the efficacy of BiDO against different attacks in the white-box setting. Note that, here we do not focus on the black-box attack methods~\cite{updates_leak,Knowledge_Align}, which are relatively easy to be defended~\cite{updates_leak,purifier}. The baseline that we will compare against is MID~\cite{MID}, which achieved the state-of-the-art defense results among the thread of generic defense mechanisms against MI attacks on DNNs.

\subsection{Experimental Setting}

\paragraph{\bf Datasets}
We study defense models built for different classification tasks, including face recognition, digit classification, and object classification. For face recognition, we use the \emph{CelebFaces Attributes Dataset} (CelebA)~\cite{celeba}. For digit classification, we use the MNIST handwritten digit data~\cite{MNIST}. For object classification, we use the CIFAR-10 dataset~\cite{cifar10}.

\paragraph{\bf Target Classifiers}
We adopt the same target classifiers as in~\cite{DMI,VMI,GMI}. For the face recognition task, we use VGG16 adapted from~\cite{vgg16} and ResNet-34 adapted from~\cite{resnet34}. For digit classification on MNIST, we use LeNet adapted from~\cite{MNIST}. For object classification, we use VGG16 adapted from~\cite{vgg16}.

\paragraph{\bf Privacy-Leakage Evaluation Methods}
We evaluate the efficacy of BiDO against the following white-box MI attacks, which are the most effective ones presented in the literature thus far.
\begin{itemize}
    \item \underline{\emph{Generative MI attack} (GMI)}~\cite{GMI} is the first white-box MI attack algorithm based on GAN~\cite{DCGAN} to achieve remarkable performance against DNNs. GMI solves the MAP to recover the most possible private attribute via gradient descent. The key idea of GMI is to leverage public data to learn a generic prior for the private training data distribution and use it to regularize the optimization problem underlying the attack.
    
    \item \underline{\emph{Knowledge-Enriched Distributional MI attack} (KED-MI)}~\cite{DMI} is an improvement on GMI, which addresses two key limitations in GMI. Firstly, they propose to tailor the training objective for building an inversion-specific GAN, where they leverage the target model to label the public dataset and train the discriminator to differentiate not only the real and fake samples but also the labels. Secondly, they recover the private data distribution by explicitly parameterizing it and solving the attack optimization over the distributional parameters.
    
    \item \underline{\emph{Variational MI attack} (VMI)}~\cite{VMI} provides a unified framework for analyzing existing methods from the Bayesian probability perspective. In the implementation, the author leverage the more powerful StyleGAN~\cite{styleGAN} as the generator which allows for fine-grained control. They also leverage the more powerful deep flow models~\cite{glow} to approximate the extended latent space of a StyleGAN. This implementation can leverage the hierarchy of learned representations, and perform targeted attacks while preserving diversity.
    
\end{itemize}

\paragraph{\bf Evaluation Protocol}
\label{pg:EP}
We evaluate the performance of a defense mechanism based on the privacy-utility tradeoff. MID and BiDO have some hyper-parameters which we can tune to vary the model robustness and model performance. For BiDO, we can vary the balancing hyper-parameters $\lambda_{x}$ and $\lambda_{y}$ in Eq.~\eqref{eq:obj}; for MID, we can tune the weight parameter $\lambda$. When preventing privacy leakage on CelebA, for GMI (or KED-MI) and VMI, we let the adversary attack the same identity 5 times and 100 times, respectively, and average the results of each attack to calculate attack accuracy; besides, note that all recovered images from each attack are used to calculate the FID value. When preventing privacy leakage on MNIST and CIFAR-10, we let the adversary attack the same class for 100 times, We mitigate the instability caused by the small number of classes by conducting 5 individual attacking trials and average the results.

\paragraph{\bf Evaluation Metrics}
Evaluating the MI attack performance requires gauging the amount of private information about a target label leaked through the synthesized images. We conduct quantitative evaluations as well as qualitative evaluations through visual inspection. The following shows the quantitative metrics that we use to evaluate the attack performance.

\begin{itemize}
    \item \underline{\emph{Attack accuracy} (Attack Acc)}. To measure the performance of an attack, we build an "evaluation classifier" to predict the identities of the reconstructed images. This metrics measure how the generated samples resemble the target class. If the evaluation classifier achieves high accuracy, the attack is then thought to be successful. To ensure a fair and informative evaluation, the evaluation classifier should be as accurate as possible. 
    
    \item \underline{\emph{Fréchet inception distance} (FID)}. We employ the widely used FID to measure both the quality and diversity of the reconstructed images, which reveals how much more detailed information is leaked from the reconstructed images. FID measures the similarity between fake and real images in the embedding space given by the features of a convolutional neural network (the evaluation classifiers in the defense task). More specifically, it computes the differences between the estimated means and covariances assuming a multivariate normal distribution on the features. 

\end{itemize}

\begin{figure*}[t]
  \includegraphics[width=0.9\textwidth]{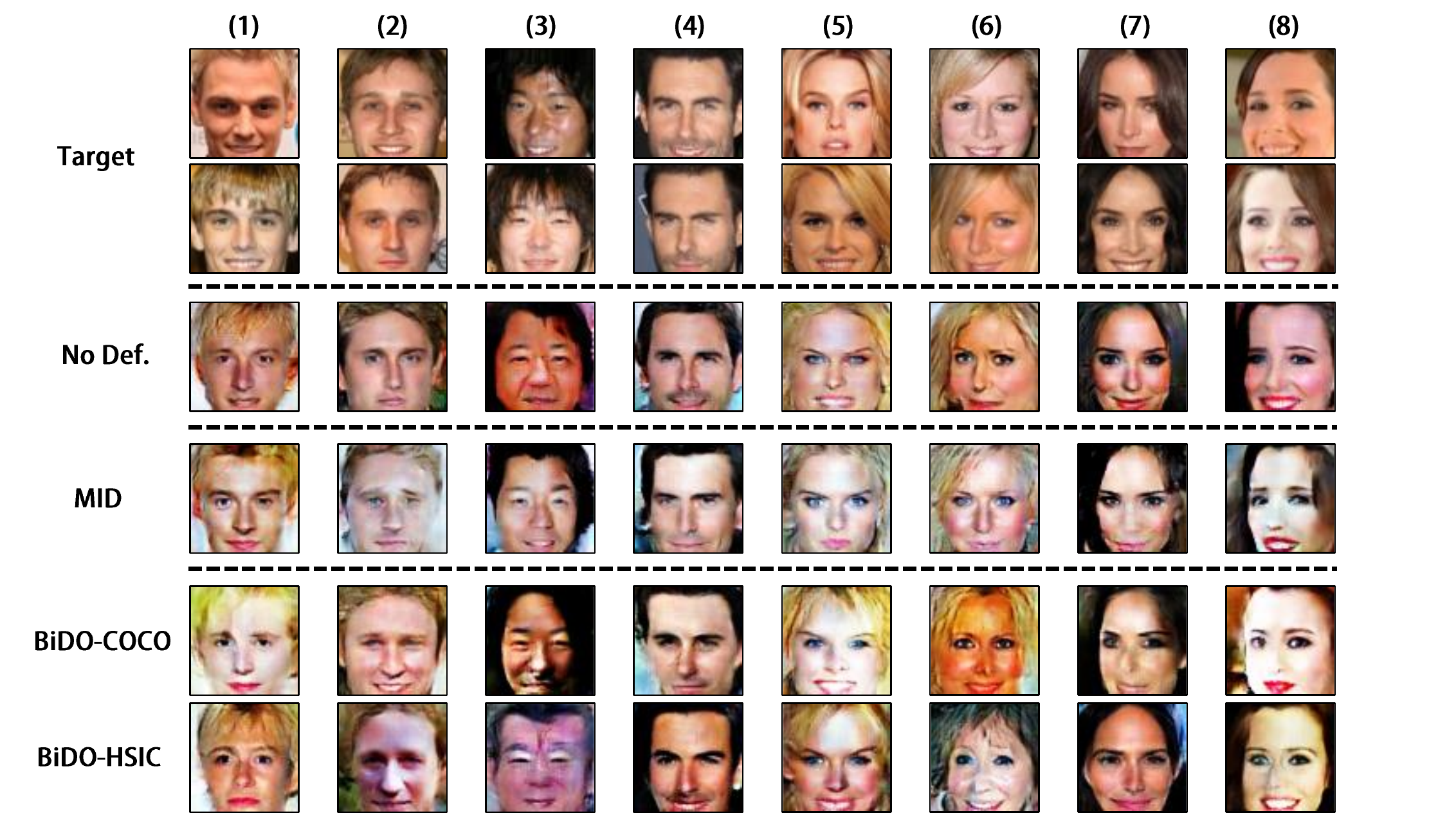}
  \vspace{-10pt}
  \caption{Qualitative comparison for preventing privacy leakage from a face recognition model trained with CelebA dataset. The first two rows show ground-truth images for target identities, images in a column belong to the same person; the third row shows reconstructed images by inverting a well-trained classifier with no defense; the fourth and last two rows illustrate the defense results of MID and BiDO, respectively.}
  \label{fig:face}
\end{figure*}

\subsection{Results}
\paragraph{\bf Performance Evaluation on CelebA}
Since facial recognition models are widely used in real scenarios, in this section, we mainly evaluate the efficacy of MID and BiDO for preventing privacy leakage on CelebA against various white-box attacks~\cite{GMI,DMI,VMI}. 

The defense results against GMI are presented in Table~\ref{tbl:GMI_celeba}, where we have models with varying model robustness and model utility by tuning the hyper-parameters. As shown in Table~\ref{tbl:GMI_celeba}, BiDO consistently achieves better utility-privacy trade-off than MID, it is noteworthy that BiDO-HSIC (c) outperforms MID (c) by a large margin in terms of attack accuracy (6.47 vs. 15.73), top-5 attack accuracy (16.07 vs. 35.27) and FID (143.63 vs. 126.24) with even better model performance (80.35 vs. 78.70). When compared with the classifier with no defense, BiDO-HSIC (c) significantly reduces the attack accuracy for more than $10\%$ (6.47 vs. 16.73) and top-5 attack accuracy for $\sim 20\%$ (35.93 vs. 16.07) while only suffering $\sim 6\%$ classification accuracy decline (80.35 vs. 86.21).

\begin{table}[!t]
  \centering
  \setlength{\belowcaptionskip}{2mm}
  \caption{Privacy leakage indicated by Attack Acc/Attack Acc5$\pm$standard deviation ($\%$) and FID on CelebA against GMI. For the MID and BiDO, we vary the weight parameter $\lambda$ and $(\lambda_{x}, \lambda_{y})$ in Eq.~\eqref{eq:obj} respectively to trade-off between model robustness and model utility. Bold values represent the setting that achieves the best trade-off between model robustness and model utility. The notion $\downarrow$ ($\uparrow$) indicates smaller (larger) values are preferred. See Section~\ref{pg:EP} for evaluation details.}
  \label{tbl:GMI_celeba}
  \resizebox{0.48\textwidth}{!}{
    \begin{tabular}{c c c c c c}
    \toprule
                     &Varying settings &Accuracy~$\uparrow$        & Attack Acc~$\downarrow$ & Attack Acc5~$\downarrow$  & FID~$\uparrow$ \\\midrule
                            
                     No Def.  & - &86.21                        & 16.73$\pm$3.32            & 35.93$\pm$4.88    & 128.81   \\\midrule
    
                            \multirow{3}{*}{MID}  & MID (a)&53.94    & 8.13$\pm$2.21        & 20.33$\pm$3.16   & 139.75   \\
                                                  & MID (b)&68.39    & 12.40$\pm$3.63       & 27.93$\pm$4.04   & 133.04 \\     
                                                  & MID (c)&78.70    & 15.73$\pm$4.11       & 35.27$\pm$4.52    & 126.24 \\\midrule
    
                      \multirow{3}{*}{BiDO-COCO}  & BiDO-COCO (a)&53.39  & 7.40$\pm$2.01    & 19.80$\pm$4.21      & 148.68 \\
                                                  & BiDO-COCO (b)&74.53  & 11.00$\pm$3.38   & 28.80$\pm$5.10       & 136.49 \\     
                                                  & BiDO-COCO (c)&81.55  & 13.67$\pm$3.49   & 32.07$\pm$3.41      & 133.92 \\\midrule

                    \multirow{3}{*}{BiDO-HSIC} & BiDO-HSIC (a)&53.49     & 1.67$\pm$1.04      & 4.93$\pm$1.64     & 180.07   \\
                                               & BiDO-HSIC (b)&70.31     & 3.13$\pm$1.69      & 9.27$\pm$4.13     & 136.49   \\
                                               & BiDO-HSIC (c)&\textbf{80.35}    & \textbf{6.47$\pm$2.57}       & \textbf{16.07$\pm$2.91}    & \textbf{134.63}   \\\bottomrule
\end{tabular}
}
\vspace{-10pt}
\end{table}

The defense results against KED-MI are illustrated in Table~\ref{tbl:DMI_celeba} where we use the same parameter settings as in defending against GMI. Except the case that BiDO-COCO (a) performs slightly worse than MID (a), on the whole BiDO could be more preferable due to its broad applicability. BiDO-HSIC can significantly prevent privacy leakage, as a highlight, BiDO-HSIC (c) outperforms MID (c) by a large margin in terms of attack accuracy (46.53 vs. 69.60), top-5 attack accuracy (73.00 vs. 92.00) and FID (250.88 vs. 235.61) with even better model performance (80.35 vs. 78.70). When compared with the classifier with no defense, BiDO-HSIC (c) reduces attack accuracy for more than $29\%$ (46.53 vs. 75.93), top-5 attack accuracy for more than $20\%$ (73.00 vs.93.80) while only suffering $\sim 6\%$ classification accuracy decline (80.35 vs. 86.21).

We also verify the performance of BiDO in Figure~\ref{fig:face}. Since KED-MI can recover the distribution w.r.t. a specific identity, we select two images from the training set that are most similar to the recovered ones. The target classifiers trained with BiDO and MID have comparable performance. We can see that BiDO can block attack much better than MID. For instance, the reconstructions for MID can still retain many facial features of the target individuals (e.g., columns $1,2,3$), while the reconstructions for the classifier trained with BiDO are either with severely damaged facial features (e.g., columns $7,8$ for BiDO-COCO and columns $5,8$ for BiDO-HSIC) or completely different from the target individuals (e.g., columns $1,3$ for BiDO-COCO and columns $1,6$ for BiDO-HSIC).

\begin{table}[!t]
  \centering
  \setlength{\belowcaptionskip}{2mm}
  \caption{Privacy leakage indicated by Attack Acc/Attack Acc5$\pm$standard deviation ($\%$) and FID on CelebA against KED-MI. For the MID and BiDO, we vary the weight parameter $\lambda$ and $(\lambda_{x}, \lambda_{y})$ in Eq.~\eqref{eq:obj} respectively to trade-off between model robustness and model utility. Bold values represent the setting that achieves the best trade-off between model robustness and model utility. The notion $\downarrow$ ($\uparrow$) indicates smaller (larger) values are preferred. See Section~\ref{pg:EP} for details.} 
  \label{tbl:DMI_celeba}

\resizebox{0.48\textwidth}{!}{
    \begin{tabular}{c c c c c c }
    \toprule
                            &Varying settings &Accuracy~$\uparrow$      & Attack Acc~$\downarrow$ & Attack Acc5~$\downarrow$  & FID~$\uparrow$ \\
    \midrule                        
              No Def.       & - &86.21         & 75.93$\pm$3.65          & 93.80$\pm$2.51    & 213.43   \\

    \midrule 
      \multirow{3}{*}{MID}  
                            & MID (a)&53.94         & 32.67$\pm$3.28          & 63.20$\pm$2.15    & 290.96   \\
                            
                            & MID (b)&68.39         & 54.73$\pm$1.99           & 79.67$\pm$2.62    & 278.85 \\
                            
                            & MID (c)&78.70        & 69.60$\pm$4.70           & 92.00$\pm$1.59    & 235.61 \\
    \midrule
      \multirow{3}{*}{BiDO-COCO} 
                            
                            & BiDO-COCO (a)&53.39         & 37.00$\pm$3.69            &68.20$\pm$2.73     & 286.70 \\
                            
                            & BiDO-COCO (b)&74.53         & 48.73$\pm$4.61           & 75.53$\pm$4.13    & 266.28 \\
                            
                            & BiDO-COCO (c)&81.55          & 62.93$\pm$4.72           & 88.27$\pm$2.37    & 225.70 \\
                                                  
    \midrule                                
      \multirow{3}{*}{BiDO-HSIC} 
                                  & BiDO-HSIC (a)&53.49         & 16.87$\pm$3.47          & 35.73$\pm$2.52    & 299.04   \\
                                
                                & BiDO-HSIC (b)&70.31         & 31.00$\pm$3.92          & 60.00$\pm$3.85    & 290.95   \\
                                
                                 & BiDO-HSIC (c)&\textbf{80.35}         & \textbf{46.53$\pm$3.67}          & \textbf{73.00$\pm$4.01}    & \textbf{250.88}   \\
    \bottomrule
    \end{tabular}
}
\vspace{-10pt}
\end{table}%

\begin{table*}[t!]\renewcommand{\arraystretch}{1.2}
  \centering
  \setlength{\belowcaptionskip}{2mm}
  \caption{Privacy leakage indicated by Attack Acc/Attack Acc5$\pm$standard deviation ($\%$) and FID$\pm$standard deviation on MNIST and CIFAR-10 against KED-MI. Bold values represent the setting that achieves the best trade-off between model robustness and model utility. The notion $\downarrow$ ($\uparrow$) indicates smaller (larger) values are preferred. See Section~\ref{pg:EP} for details.} 
  \label{tbl:DMI_mnist_cifar}%
\resizebox{0.9\textwidth}{!}{

\begin{tabular}{c | c c c c | c c c c}
\hline
    & \multicolumn{4}{c}{MNIST} & \multicolumn{4}{|c}{CIFAR-10} \\ \hline
   & Accuracy~$\uparrow$     & Attack Acc~$\downarrow$    & Attack Acc5~$\downarrow$  & FID~$\uparrow$ & Accuracy~$\uparrow$ & Attack Acc~$\downarrow$ & Attack Acc5~$\downarrow$  & FID~$\uparrow$\\ \hline

  No Def.  & 99.94         & 42.52$\pm$10.09         & 95.24$\pm$5.24    & 245.64$\pm$14.52   & 96.17   & 72.16$\pm$17.7           & 99.56$\pm$2.20    & 167.85$\pm$5.44\\  \hline

  MID           & 97.42        & 51.75$\pm$14.65          & 100.00$\pm$0.00    & 218.06$\pm$17.48  & 89.04   & 61.40$\pm$13.88          & 99.72$\pm$1.44    & 176.84$\pm$20.36\\  \hline

 BiDO-COCO    & 99.51         & 7.76$\pm$8.47          & 84.40$\pm$12.95          & 305.34$\pm$19.44  & \textbf{95.39}  & \textbf{56.20$\pm$17.79}&\textbf{ 98.92$\pm$3.63} &\textbf{ 179.51$\pm$16.25} \\  \hline

  BiDO-HSIC     & \textbf{99.61}        & \textbf{4.36$\pm$5.70}          & \textbf{76.96$\pm$12.10}    & \textbf{322.00$\pm$16.96}  & 93.79   & 58.24$\pm$18.63         & 99.08$\pm$3.99    & 171.60$\pm$12.45\\  \hline

\end{tabular}
}
\end{table*}

\begin{table}[!t]
  \centering
  \setlength{\belowcaptionskip}{2mm}
  \caption{Privacy leakage indicated by Attack Acc/Attack Acc5$\pm$standard deviation ($\%$) and FID on CelebA against VMI. Bold values represent the setting that achieves the best trade-off between model robustness and model utility. The notion $\downarrow$ ($\uparrow$) indicates smaller (larger) values are preferred. See Section~\ref{pg:EP} for details.} 
  \label{tbl:VMI_celeba}
\resizebox{0.42\textwidth}{!}{
\begin{threeparttable}
    \begin{tabular}{c c c c c c}
    \toprule
                      & Accuracy~$\uparrow$     & Attack Acc~$\downarrow$    & Attack Acc5~$\downarrow$  & FID~$\uparrow$\tnote{2} \\\midrule 
    No Def.           & 69.27         & 36.90$\pm$22.70          & 61.50$\pm$21.24    & 0.9295   \\\midrule 
    MID\tnote{1}      & 52.52         & 29.05$\pm$23.99          & 51.05$\pm$28.52    & 0.9854   \\\midrule
    BiDO-COCO         & 59.34         & 29.45$\pm$16.54          & 54.25$\pm$21.01    & 0.9705   \\\midrule                                
    BiDO-HSIC         & \textbf{61.14}  & \textbf{30.25$\pm$23.46}  & \textbf{53.35$\pm$27.65}    & \textbf{0.9665} \\
    \bottomrule
    \end{tabular}

     \begin{tablenotes}
       \footnotesize
       \item[1] MID with hyper-parameter $\lambda=0$.
       \item[2] Following the original setting, $l_2$ normalized features are used to calculate the FID value.
     \end{tablenotes}
\end{threeparttable}
}
\vspace{-10pt}
\end{table}

The defense results against VMI are shown in Table~\ref{tbl:VMI_celeba}. Note that, although we set hyper-parameter for the dependency regularization term $\lambda=0$, the classifier trained with MID still cannot achieve comparable classification accuracy as the classifier with no defense. We speculate the reason is: solving the variational approximation of mutual information requires modifying the last layer of the network, which makes it vulnerable to the data pre-processing method used in VMI. While BiDO does not suffer from such a problem.

\begin{table*}[!t]
  \centering
  \setlength{\belowcaptionskip}{2mm}
  \caption{Ablation study on BiDO. Rows (2-3)/Rows (5-6) indicate the effect of removing each component of the learning objective in Eq.~\eqref{eq:empirical_coco} (row (4))/Eq.~\eqref{eq:empirical_hsic} (row (7)). We evaluate each objective over Accuracy (\%) and privacy leakage indicated by Attack Acc/Attack Acc5$\pm$standard deviation ($\%$) and FID on CelebA against KED-MI. We set $\lambda_x$ as $10$ and $0.1$, $\lambda_y$ as $50$ and $1$, for BiDO-COCO and BiDO-HSIC, respectively. Bold values represent the setting that achieves the best trade-off between model robustness and model utility. The notion $\downarrow$ ($\uparrow$) indicates smaller (larger) values are preferred. See Section~\ref{pg:EP} for details.}
  \label{tbl:Ablation}%
  \resizebox{0.9\textwidth}{!}{
    \begin{tabular}{c c c c c c c}
    \toprule
                                            &Rows  &Objectives & Accuracy~$\uparrow$     & Attack Acc~$\downarrow$ & Attack Acc5~$\downarrow$ & FID~$\uparrow$ \\
    \midrule                        
                                            &(1)  &$\loss(\theta)$         
                                                
                                                & 85.31        & 62.73$\pm$4.19       &87.73.20$\pm$2.88             & 226.82   \\
    \midrule
               \multirow{3}{*}{BiDO-COCO}  & (2)         & $\loss(\theta) + \lambda_{x} \sum_{j=1}^{\nol} \coco(X, Z_j)$
                                                   
                                                    &85.04         & 63.40$\pm$3.19         & 87.93$\pm$3.30                & 218.10 \\
                                                    
                                            & (3)       & $\loss(\theta) -\lambda_{y} \sum_{j=1}^{\nol} \coco(Z_j, Y)$
                                                    
                                                    &75.13         & 59.00$\pm$4.74        & 80.80$\pm$3.09                 & 255.09 \\
                                                    
                                          & (4)       & $\loss(\theta) + \lambda_{x} \sum_{j=1}^{\nol} \coco(X, Z_j) -\lambda_{y} \sum_{j=1}^{\nol} \coco(Z_j, Y)$
                                                    
                                                    &\textbf{74.53}         & \textbf{48.73$\pm$4.61}           & \textbf{75.53$\pm$4.13}    & \textbf{266.28} \\

    \midrule                                
                \multirow{3}{*}{BiDO-HSIC} & (5)& $\loss(\theta) + \lambda_{x} \sum_{j=1}^{\nol} \hsic(X, Z_j)$
                                                
                                                &0.83             & 0.67$\pm$0.37         & 3.00$\pm$1.23                 & 720.35   \\
                                                
                                           & (6)     & $\loss(\theta) -\lambda_{y} \sum_{j=1}^{\nol} \hsic(Z_j, Y)$
                                                
                                                &64.73          & 27.47$\pm$4.80          & 52.53$\pm$4.78                & 289.21    \\
                                                
                                           & (7)     & $\loss(\theta) + \lambda_{x} \sum_{j=1}^{\nol} \hsic(X, Z_j) -\lambda_{y} \sum_{j=1}^{\nol} \hsic(Z_j, Y)$
                                                
                                                &\textbf{76.36} & \textbf{31.40$\pm$4.38}    &\textbf{55.20$\pm$3.09}   &\textbf{276.21}   \\
    \bottomrule
    \end{tabular}
    }
\end{table*}%

\vspace{-0.5em}\paragraph{\bf Performance Evaluation on MNIST and CIFAR-10}

In this section, we evaluate privacy leakage on MNIST and CIFAR-10 against KED-MI. Table~\ref{tbl:DMI_mnist_cifar} shows the defense results, where BiDO achieves superior defense performance than MID on both datasets. Take results on MNIST for example, the classifier trained with MID exposes more information about the training data in terms of attack accuracy (51.75 vs. 42.52) than the classifier with no defense, even with worse model performance (97.42 vs. 99.94), while BiDO effectively improves the model's robustness with a slight decline in classification accuracy. For example, BiDO-HSIC remarkably reduces attack accuracy by $\sim38\%$ (4.36 vs. 42.52) with only $\sim0.3\%$ classification accuracy decline (99.61 vs. 99.94).

\begin{figure}[!t]
  \includegraphics[width=0.43\textwidth]{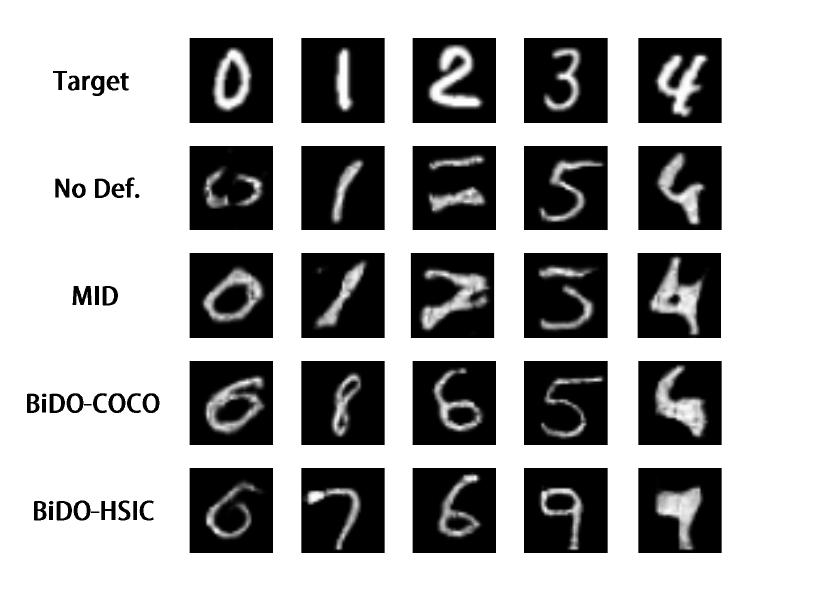}
  \vspace{-10pt}
  \caption{Qualitative comparison for preventing privacy leakage from a digit classification model trained with MNIST. The first row shows ground-truth images for target digits; the second row shows reconstructed images by inverting a well-trained classifier with no defense; the third and last two rows illustrate the results of MID and BiDO, respectively.}
  \label{fig:digit}
  \vspace{-10 pt}
\end{figure}

The qualitative defense results against KED-MI are presented in Figure~\ref{fig:digit}, we can see that BiDO greatly prevents information leakage, the recovered digits for the classifier trained with BiDO are virtually different from the target ones; while for MID, the recovered images may leak more information about the target digits (e.g., digits $2,3$).

\paragraph{\bf Ablation Study}
In this section, we turn our attention to how the last two terms in the loss function in Eq.~\eqref{eq:empirical_coco} and Eq.~\eqref{eq:empirical_hsic} affect model robustness against MI attacks. As illustrated in Table~\ref{tbl:Ablation}, removing any parts leads to a worse privacy-utility trade-off. Specifically, for BiDO-COCO, removing the  $\coco(Z_j,Y)$ penalty (row (2)) brings little improvement in model robustness, because the value of $d(X,Z_j)$ is too small to affect the training process; while for BiDO-HSIC, removing the penalty on $\hsic(Z_j,Y)$ (row (5)) degrades the classification accuracy significantly, which means coupling learning to labels via the third term is integral to maintaining useful label-related information in latent layers; for BiDO-COCO, removing $\coco(X,Z_j)$ penalty (row (3)) degrades defense performance since the dependency between inputs and latent representations makes the model vulnerable to MI attacks; while for BiDO-HSIC, without $\hsic(X,Z_j)$ penalty (row (6)), the model is overfitted on the training set, thus the classification accuracy decreased. The two dependency terms combined by proper hyper-parameters $\lambda_x$ and $\lambda_y$ (rows (4, 7)) achieve a better privacy-utility trade-off.

\section{Conclusion}
For defending against \emph{model-inversion} (MI) attacks, we propose a \emph{bilateral dependency optimization} (BiDO) strategy, which aims at minimizing the dependency between the latent representations and the inputs while maximizing the dependency between latent representations and the outputs. In particular, the BiDO strategy is achieved by using the dependency constraints as a universally applicable regularizer in addition to commonly used losses for deep neural networks, which can be instantiated with appropriate dependency criteria according to different tasks. We verified the efficacy of BiDO by using two different dependency measures: \emph{BiDO with constrained covariance} (BiDO-COCO) and \emph{BiDO with Hilbert-Schmidt Independence Criterion} (BiDO-HSIC). Experiments show that BiDO achieves the state-of-the-art defense performance for a variety of datasets, target models and MI attacks, which lights up a novel road to defend against MI attacks.

\section*{Acknowledgments and Disclosure of Funding}
XP and BH were supported by the RGC Early Career Scheme No. 22200720, NSFC Young Scientists Fund No. 62006202, Guangdong Basic and Applied Basic Research Foundation No. 2022A1515011652, RIKEN Collaborative Research Fund and HKBU CSD Departmental Incentive Grant. JFZ was supported by JSPS Grants-in-Aid for Scientific Research (KAKENHI), Early-Career Scientists, Grant Number 22K17955, Japan JST Strategic Basic Research Programs, ACT-X, Grant Number JPMJAX21AF, Japan.


\bibliographystyle{ACM-Reference-Format}
\bibliography{refs}

\newpage    

\appendix

\section{Data Links and Code}
Data links and code can be found in this repository: {\url{https://github.com/xpeng9719/Defend_MI}}.

\section{Experimental Details}
\label{Asec:ed}
\subsection{Datasets and Evaluation Classifiers}
For defending against different MI attacks, we adopt the same dataset settings and evaluation classifiers as the original papers. Please refer to GMI~\cite{GMI}, KED-MI~\cite{DMI} and VMI~\cite{VMI} for more details.

\subsection{Training Configurations of Target Classifiers}
\paragraph{\bf CelebA}
For GMI and KED-MI, the target classifier for CelebA was a VGG16, a pretrained checkpoint from Pytorch Hub (\url{https://pytorch.org/hub/}) is utilized to initialize parameters of the classifier. It was trained using Adam~\cite{adam} (learning rate=1e-4, batch size=64) for 50 epochs. For VMI, the target classifier was a ResNet-34. It was trained using SGD with Nestrov momentum (learning rate=1e-1, batch size=64, momentum=0.9, weight decay=5e-4) for 100 epochs. Learning rate decayed by a factor of 0.2 at 60 and 80 epochs.

\paragraph{\bf MNIST}
For KED-MI, the target classifer for MNIST is a adapted LeNet. It is trained using Adam~\cite{adam} (learning rate=1e-1, batch size=64) for 20 epochs.

\paragraph{\bf CIFAR-10}
The target classifier for CIFAR-10 was a VGG16, a pretrained checkpoint from Pytorch Hub is utilized to initialize parameters of the classifier. It was trained using Adam~\cite{adam} (learning rate=1e-4, batch size=64) for 30 epochs. Learning rate decayed by a factor of 0.5 at 25 epochs.

\begin{table}[h]
  \centering
  \caption{Hyper-parameter settings on CelebA against GMI and KED-MI: For the MID and BiDO, the balancing hyper-parameters are $\lambda$ and $(\lambda_{x}, \lambda_{y})$ in Eq.~\eqref{eq:obj}, respectively.} 
  \label{tbl:hp_celeba_GD}
  \resizebox{0.35\textwidth}{!}{
    \begin{tabular}{c c c c c c}
    \toprule
                     &Varying settings &Hyper-parameters    \\\midrule
                            
        \multirow{3}{*}{MID}  & MID (a)&0.02    \\
                              & MID (b)&0.01     \\     
                              & MID (c)&0.003     \\\midrule
    
        \multirow{3}{*}{BiDO-COCO}  & BiDO-COCO (a)&(15, 75)   \\
                                    & BiDO-COCO (b)&(10, 50)   \\     
                                    & BiDO-COCO (c)&(5, 50)   \\\midrule

        \multirow{3}{*}{BiDO-HSIC} & BiDO-HSIC (a)&(0.05, 2.5)       \\
                                   & BiDO-HSIC (b)&(0.05, 1)     \\
                                   & BiDO-HSIC (c)&(0.05, 0.5)    \\\bottomrule
\end{tabular}
}
\end{table}

\begin{table}[h]
  \centering
  \caption{Hyper-parameter settings on CelebA against VMI: For the MID and BiDO, the balancing hyper-parameters are $\lambda$ and $(\lambda_{x}, \lambda_{y})$ in Eq.~\eqref{eq:obj}, respectively.} 
  \label{tbl:hp_celeba_VMI}
  \resizebox{0.28\textwidth}{!}{
    \begin{tabular}{c | c c c c c}
    \hline
                      &Hyper-parameters    \\\hline
                            
        \multirow{1}{*}{MID}   & 0     \\\hline
    
        \multirow{1}{*}{BiDO-COCO} &(0.05, 2.5)   \\\hline

        \multirow{1}{*}{BiDO-HSIC} &(0.1, 2)    \\\hline
\end{tabular}
}
\end{table}
\begin{table}[h]\renewcommand{\arraystretch}{1.2}
  \centering
  \caption{Hyper-parameter settings on MNIST and CIFAR-10 against KED-MI: For the MID and BiDO, the balancing hyper-parameters are $\lambda$ and $(\lambda_{x}, \lambda_{y})$ in Eq.~\eqref{eq:obj}, respectively.} 
  \label{tbl:hp_DMI_MC}%
\resizebox{0.4\textwidth}{!}{

\begin{tabular}{c | c |  c}
\hline
            & MNIST &  CIFAR-10 \\ \hline
            & Hyper-parameters      & Hyper-parameters \\ \hline
 
  MID       & 0.2       & 0.15 \\  \hline

 BiDO-COCO  & (1, 50)   & (0.1, 5) \\  \hline
                                
 BiDO-HSIC  & (2, 20)  & (0.1, 1)    \\\hline

\end{tabular}
}
\end{table}

\section{Algorithm Details and Balancing Hyper-parameter Settings}
\label{appendix:hp_setting}
In both BiDO-COCO and BiDO-HSIC, we apply Gaussian kernels for inputs $X$ and latent representations $Z$, and a linear kernel for labels $Y$. For Gaussian kernels, we set $\sigma=5\sqrt{d}$, where $d$ is the dimension of the corresponding random variable.

Hyper-parameter settings used in Table~\ref{tbl:GMI_celeba} and Table~\ref{tbl:DMI_celeba} are reported in Table~\ref{tbl:hp_celeba_GD}. When we conduct evaluation on CelebA against GMI and KED-MI, for BiDO-COCO, we have tested $\lambda_x \in [1, 15], \frac{\lambda_y}{\lambda_x} \in [5, 50]$; for BiDO-HSIC, we have tested $\lambda_x \in [0.01, 0.2], \frac{\lambda_y}{\lambda_x} \in [5, 50]$. Hyper-parameter settings used in Table~\ref{tbl:VMI_celeba} are reported in Table~\ref{tbl:hp_celeba_VMI}. When we conduct evaluation on CelebA against VMI, for both BiDO-COCO and BiDO-HSIC, we have tested $\lambda_x \in [0.05, 1], \frac{\lambda_y}{\lambda_x} \in [5, 50]$. Typically larger $\frac{\lambda_y}{\lambda_x}$ or larger hyper-parameter values will lead to lower predictive power of the classifier and thus lower attack accuracy of the attack algorithm.

Hyper-parameter settings used in Table~\ref{tbl:DMI_mnist_cifar} are reported in Table~\ref{tbl:hp_DMI_MC}. When we conduct evaluation on MNIST and CIFAR-10 against KED-MI, for both BiDO-COCO and BiDO-HSIC, we have tested $\lambda_x \in [0.1, 2], \frac{\lambda_y}{\lambda_x} \in [5, 50]$. Typically larger $\frac{\lambda_y}{\lambda_x}$ or larger hyper-parameter values will lead to lower predictive power of the classifier and thus lower attack accuracy of the attack algorithm.

\end{document}